\documentclass[sigconf]{acmart}
\usepackage{graphicx}
\usepackage{array}  
\usepackage{multirow} 
\graphicspath{{figures/}}
\usepackage{CJKutf8}

\newcommand{\RNum}[1]{\uppercase\expandafter{\romannumeral #1\relax}}
\AtBeginDocument{%
	\providecommand\BibTeX{{%
			\normalfont B\kern-0.5em{\scshape i\kern-0.25em b}\kern-0.8em\TeX}}}

\setcopyright{acmcopyright}
\copyrightyear{2018}
\acmYear{2018}
\acmDOI{10.1145/1122445.1122456}

\acmConference[Woodstock '18]{Woodstock '18: ACM Symposium on Neural
	Gaze Detection}{June 03--05, 2018}{Woodstock, NY}
\acmBooktitle{Woodstock '18: ACM Symposium on Neural Gaze Detection,
	June 03--05, 2018, Woodstock, NY}
\acmPrice{15.00}
\acmISBN{978-1-4503-XXXX-X/18/06}

\begin{document}
	\begin{CJK*}{UTF8}{gkai} 
		\title{Multi-Fusion Chinese WordNet (MCW) : Compound of Machine Learning and Manual Correction}
		
		%
		%
		%
		%
		%
		%
		\author{Mingchen Li}
		\affiliation{\institution{Qufu Normal University}}
		\email{lmingchen96@gmail.com}
		\author{ Zili Zhou}
		\affiliation{\institution{Qufu Normal University}}
		\email{zlzhou222@163.com}
		
		\author{Yanna Wang}
		\affiliation{\institution{Qufu Normal University}}
		\email{yannawang2008@163.com}

		%
		
		\begin{abstract}
			Princeton WordNet (PWN) is a lexicon-semantic network based on cognitive linguistics, which promotes the development of natural language processing. Based on PWN, five Chinese wordnets have been developed to solve the problems of syntax and semantics. They include: Northeastern University Chinese WordNet (NEW), Sinica Bilingual Ontological WordNet (BOW), Southeast University Chinese WordNet (SEW), Taiwan University Chinese WordNet (CWN), Chinese Open WordNet (COW). By using them, we found that these word networks have low accuracy and coverage, and cannot completely portray the semantic network of PWN. So we decided to make a new Chinese wordnet called Multi-Fusion Chinese Wordnet (MCW) to make up those shortcomings. The key idea is to extend the SEW with the help of Oxford bilingual dictionary and Xinhua bilingual dictionary, and then correct it. More specifically, we used machine learning and manual adjustment in our corrections. Two standards were formulated to help our work. We conducted experiments on three tasks including relatedness calculation, word similarity and word sense disambiguation for the comparison of lemma's accuracy, at the same time, coverage also was compared. The results indicate that MCW can benefit from coverage and accuracy via our method. However, it still has room for improvement, especially with lemmas. In the future, we will continue to enhance the accuracy of MCW and expand the concepts in it. 
			
		\end{abstract}

		
		
		
		\maketitle

	\section{Introduction}
	A multi-level, multi-type, multi-relational semantic system will help accomplish different natural language tasks. PWN's \cite{Miller1995WordNet} proposal promotes the development of this system. It has been successfully applied to Word Sense Disambiguation \cite{Bradford2008Word}, Machine Translation \cite{Mikolov2013Exploiting}, and a series of language projects. In academia, it was widely regarded as the most important resource for computational linguistics \cite{Sproaty1999Computational}. Many countries have already started to build national wordnet. The Netherlands, Spain, Britain, France and some other countries participated in the construction of Europe wordnet (EuroNet) \cite{Vossen1998EuroWordNet}. Korea also built Korean wordnet (KoreaNet) \cite{Lee2000Automatic}. Based on PWN, five Chinese wordnets also have been developed. In 2003, Northeastern University Chinese WordNet (NEW) \cite{ZhangLi2003Chinese} was created, this is a Chinese wordnet transformation generating system. In 2004, Sinica Bilingual Ontological WordNet (BOW) \cite{Huang2004Sinica} was created through a bootstrapping method. In 2008 Southeast University Chinese WordNet(SEW) \cite{Xu2008An} was automatically constructed by implementing three approaches, including Minimum Distance (MDA), Intersection (IA) and Words Co-occurrence (WCA). In 2010, Taiwan University and Academia Sinica constructed the Chinese WordNet (CWN) \cite{Huang2010Chinese}, using a method of combining the analysis and corpus. In 2013, Nanyang Technological University constructed the Chinese Open WordNet (COW) \cite{Wang2014Building} by manually, it combines the SEW and Wiktionary (WIKT).
	
	Researchers have been made some contributions to the Chinese wordnet, however, all of these wordnets have some flaws, they don't have the enough lemmas and high  accuracy, especially with polysemous words, some are constructed by traditional Chinese, it is inconvenient to use. A high quality Chinese wordnet would be an important source for the community. So, we started to create Multi-Fusion Chinese WordNet (MCW). The experiments show that the accuracy and the coverage performance have an obvious improvement. Our work is based on wordnet 3.0, and meets the PWN's build standard.
	
	In this paper, we pioneered three tasks to evaluate the Chinese wordnets, due to lack of the licence of BOW and NEW, we only compared the other four Chinese wordnets (COW, SEW, CWN, MCW) and showed their strengths and weaknesses by experiments.
	
	The following sections are organized as follows. Section 2 elaborates on the construction process of the three Chinese wordnets. Section 3 introduces our work in building MCW. Section 4 compares the four  Chinese wordnets in different aspects. Finally the conclusion and future work are stated in Section 5.

	\section{Related Research}
	In 1985, a group of psychological lexicologists and linguists at Princeton University began to develop a dictionary database \cite{Miller1990Introduction}, called PWN. Nouns, verbs, adjectives and adverbs are grouped into sets of cognitive synonyms (synsets) in PWN. PWN established a variety of lexical and semantic relations among these concepts, including synonymy, antisense, hypernym and hyponym \cite{Miller1995WordNet,Miller1990Introduction}. It is considered to be the most important resource in many fields, such as computational linguistics, and psycholinguistics \cite{Gernsbacher1994Handbook}. Hence many countries have built wordnet by their own language based on PWN. Following the trend, Chinese WordNets are also constructed. We will describe their construction method in the next, they inculde SEW,CWN and COW.
	
	SEW is a bilingual lexical database which in simplified Chinese and based on wordnet2.0. It consists of more than 110,000 concepts, over 150,000 words. In the progress of constructing, it implements three methods, including Minimum Distance (MDA), Intersection (IA) and Words Co-occurrence (WCA). MDA calculates the Levenshtein Distance \cite{Miller2009Levenshtein} between gloss of synset and explanations in American Heritage Dictionary. IA can find a proper translation, even synset contains only one word in X-Dict Dictionary. WCA follows the principle that the co-occurrence \cite{Y2004KEYWORD} frequency which refers to the relation between these words. IA gets the highest precision, but lowest recall. WCA covers 57.2\% synsets, it has the widest processing ability, but also the lowest precision. Synthesize these methods, the MIWA was proposed, which integrates three approaches. It covers 70\% of synsets and got 61.5\% F-measure.

	The design concept of CWN is to take into account the precise expression of lemmas and lemmas' relations under a complete knowledge system. Its translation based on wordnet1.6. It used the upper layer shared knowledge ontology (SUMO) \cite{Pease2002The} to provide a standardized system representation of knowledge. CWN is in traditional Chinese, the vocabulary expression is slightly different from the simplified Chinese.
	
	COW was built in 2013 which based on wordnet3.0. It used SEW to sense-tag a corpus, in addition, a small scale Chinese wordnet was constructed from open multilingual wordnet (OMW) \cite{Bond2013Linking} using data from WIKI. Before construction, they made a revision of the PWN, there are 1,840 wrong entities (15\%) in the PWN, they deleted 1,706 translations and amended 134. Furthermore, they added 2604 new entities (about 21\%) and checked the wrong entities according to POS. The highlight is the manual correction, although there is no full lemmas of COW, the accuracy was improved. Its F-score is 81\%, above other Chinese wordnets.
	
	Besides the above Chinese wordnets, we created a novel Chinese wordnet (MCW) and used the Oxford bilingual dictionary, Xinhua bilingual dictionary to aid our work. Those dictionaries can help us improve coverage. In addition, we used machine learning and manual correction to improve accuracy.
	
	For the sake of contrastive analysis, we did some cleaning up and wordnets (1.6, 2.0) were mapped into wordnet3.0, traditional Chinese was converted into simplified Chinese. The map list is available at \url {https://github.com/ToneLi/wordnet_mapid}
	
	\section{Build the  Multi-Fusion Chinese WordNet }
	
	SEW has 95.78\% concepts, it can help us construct MCW, but this is still a shortage of complete concept compared with wordnet 3.0, so we used two dictionaries to expend it. Machine learning can help with initial screening, manual correction is the guarantee to improve accuracy. Our approach includes the following three stages: Based on SEW, expand the number of lemmas; Primary screening using machine learning \cite{Quinlan1992C4}; Secondary screening by manual correction. MCW has the same license as the PWN. Translated version is wordnet3.0 and content is simplified Chinese.

	\subsection{Expand lemmas}
	
	We have used a variety of bilingual dictionaries, this includes: Oxford bilingual dictionary and Xinhua bilingual dictionary. Lemma in PWN can find its translation in these dictionaries. But each concept has only one meaning in PWN, directly translated lemmas by these dictionaries can make some concepts have more than one meanings, it was shown in Figure 1.

	\begin{figure}[htbp]  
		\centering
		\includegraphics[height=3.5cm,width=0.28\textheight]{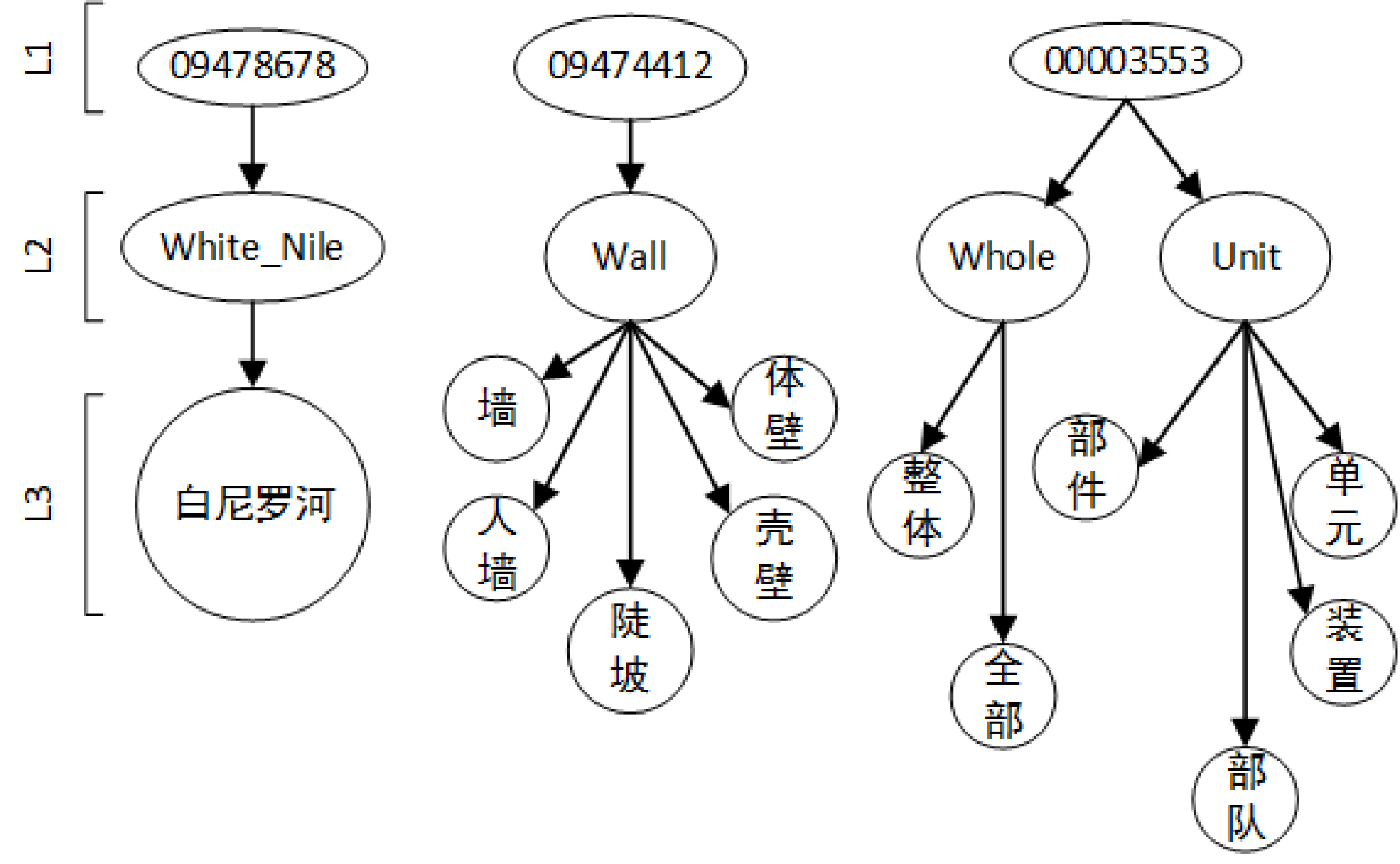}
		\caption{L1 means the ids of concepts in PWN, L2 means the lemmas in every concept, L3 means the Chinese lemmas translated by dictionary.}\label{1} 
	\end{figure}
	
	From Fig.1, a concept has more than one meaning,  such as: in 00003553, 整体-zhengti (an integral part of the whole) and 部队-budui (a general term for an army) are different meanings in Chinese.
	
	We divided the concepts into 3 categories after translation. Category 1: Every concept in PWN has one lemma (one meaning), also in Chinese, such as: in 09478678, it has one lemma (White\_Nile), one Chinese lemma (白尼罗河-bainiluohe), both of them have one meaning. Category 2: Every concept in PWN has one lemma, but different meanings. more than one lemma in Chinese, also different meanings.
	such as: in 09474412,  it has one lemma (Wall), different lemmas in Chinese, those different lemmas have different meanings, 体壁-tibi and 陡坡-doupo have diferent meanings. Category 3: Every concept in PWN has more than one lemma, different meanings, more than one lemma in Chinese, also different meanings. (these examples are from Fig.1)
	
	Classification is the basis of preliminary screening. After Classification, we merged these lemmas with SEW. This combination contains 117659 concepts and 265432 lemmas.

	\subsection{Preliminary screening}
	This section will describe how to filer the synset lemmas in every concept by machine learning.
	
	We got each Chinese lemma's vector by Word2vec (skip-gram) \cite{Turian2010Word,Mikolov2013Efficient,Ma2017Multi} and using Principal Component Analysis(PCA) \cite{Shlens2014A} for dimensionality reduction, then mapped them to 2D space and kept an assembly of dense points.
	\subsubsection{Model Settings}
	We used word2vec (skip-gram) as our vector calculation model. We adopt the Chinese Wikipedia Dump as our training corpus, the initial learning rate to be 0.0001, the window size to be 5, the word vector dimension to be 200. Words with frequency less than 1 were ignored during training.
	\subsubsection{Dimensionality Reduction}
	We processed 200-dimensional vectors and mapped them to 2D space. Through PCA, original information is preserved. Vector is easier to process and easier to use in low dimension.
	
	\subsubsection{Lemmas Selection}
	The formula about selection was defined, as follows:
	$$s=
	\begin{cases}
	L_i,L_j& {|L_i|-|L_j|<n \cup len(lems)>2}\\
	del& {else}
	\end{cases}  \eqno{(1)}$$
	
	In this formula, $|L_i|$ and $|L_j|$ mean the absolute value of the distance from any lemmas to the origin. For every concept, when the distance between any lemmas is less than threshold value (0.21), these lemmas are retained. The following example further exlpain: (08272961-n 结合-jiehe [a],组合-zuhe [b],联合-lainhe [c],联合体-lainheti [d]).
	\begin{figure}[htbp]  
		\centering
		\includegraphics[height=3.4cm,width=0.23\textheight]{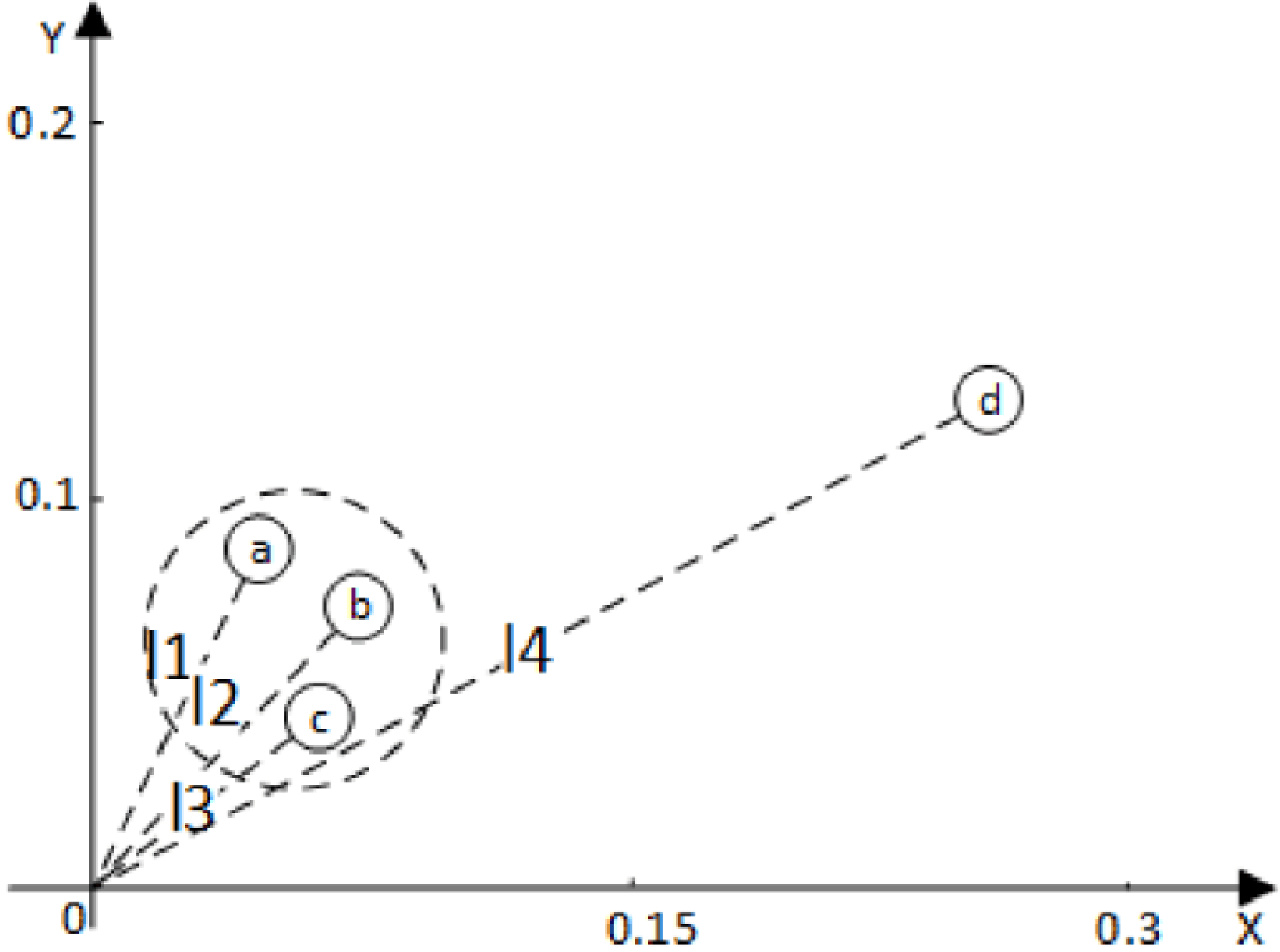}
		\caption{The vector distribution of lemmas in 08272961. X and Y represent vector value of two dimensions, (l1, l2, l3, l4) represent the absolute length of vector }\label{fig-2} 
	\end{figure}
	
	In Fig.2, the value between any two in (l1, l2, l3)  is less than a threshold value (0.21), however the value of l4 is bigger, so, (a, b, c) are reserved. The synset lemmas of 08272961 are 结合-jiehe,组合-zuhe,联合-lainhe.

		\begin{table*}[!htbp]
		\caption{Comparison of the number of concepts for Chinese wordnets. The symbol C-N means the number of concepts in each Chinese wordnet. The symbol W-(2.0,1.6,3.0) means the number of concepts in different versions of PWN. The symbol T means the percentage of Chinese wordnets translation.}
		\centering 
		\renewcommand\arraystretch{1.3}  
		\begin{tabular}{p{0.7cm}<{\centering} p{0.9cm}<{\centering} p{1.0cm}<{\centering} p{0.7cm}<{\centering} p{0.8cm}<{\centering} p{1.0cm}<{\centering} p{0.7cm}<{\centering} p{0.8cm}<{\centering} p{1.0cm}<{\centering} p{0.7cm}<{\centering} p{0.9cm}<{\centering} p{1.0cm}<{\centering} p{0.3cm}<{\centering}}
			
			\hline
			POS 
			&\multicolumn{3}{c }{SEW}
			&\multicolumn{3}{c }{CWN}
			&\multicolumn{3}{c }{COW}
			&\multicolumn{3}{c }{MCW}   \\
			
			label& C-N & W-2.0& T & C-N& W-1.6 & T& C-N& W-3.0& T& C-N& W-3.0 & T \\ 	
			\hline
			noun& 75195 &79689& 0.944& 66024& 66025 &0.999& 27888& 82115 & 0.339& 82115 &82115  & 1.0 \\ 
			verb& 13373 & 13508&0.990 & 12127& 12127 & 1.000& 5158& 13767 &0.375& 13767& 13767 & 1.0 \\ 
			adj& 18371 &18563&0.990 & 17914& 17915 & 0.999& 8559& 18156 & 0.471& 18156&18156 & 1.0 \\ 
			adv& 3618 & 3664&0.987 & 3575& 3575 & 1.000& 708& 3621&0.196& 3621&3621 & 1.0 \\ 
			Total& 110557 & 115424&0.958 & 99640& 99642 & 0.999& 42313& 117659 & 0.359&117659&117659 & 1.0 \\ 
			\hline

		\end{tabular}
	\end{table*}
	\subsection{Secondary screening}
	This section will describe how to manually correct in some wrong translations and unified the translations of concept. Finally we will list several difficulties in constructing a special vocabulary.
	\subsubsection{Correction of Wrong Translation}
	From three main aspects to correct translation.
	
	\textbf{\RNum {1}} In concept, if the lemma's translation does not reflect the meaning of the concept, delete it. The example is as follows:
	\begin{enumerate}
		\item 05960464-n dogma ( a doctrine or code of beliefs accepted as authoritative; "he believed all the Marxist dogma"), (教条-jiaotiao,教理-jiaoli,信仰-xinyang).
	\end{enumerate} 
	The lemma (信仰-xinyang) is a quality of a person, it does not match the meaning of concept 05960464, so it should be deleted.
	
	\textbf{\RNum {2}} The pos of some concepts is not right, we should correct them.
	\begin{enumerate}
		\item 00285314-v verdigris (color verdigris),(变铜绿-baintonglv).
	\end{enumerate}
	from this concept, it is a color, not verb. So the translation of it is (铜绿-tonglv).
	
	\textbf{\RNum {3}} The polysemy exists in the PWN, the machine cannot distinguish them correctly when they are translated into Chinese, we need to manually distinguish them.
	\begin{enumerate}
		\item 00047610-v wear (have or show an appearance of; "wear one's hair in a certain way"),(戴-dai,穿-chuan).
		\vspace{0.5em}
		\item 00469382-v wear (deteriorate through use or stress; "The constant friction wore out the cloth"),(穿-chuan)
	\end{enumerate}

	00047610 and 00469382 have the same translation "穿-chuan", but they do not have the same meaning, it can describe from the gloss. Wear's translation is not right in 00469382, it should be (磨损-mosun).
	
	\subsubsection{Unified the Translation}
	For ease to use, we have unified translations manually.
	
	\textbf{\RNum {1}} In English, Place names and language names sometimes can be replaced by one word, but in Chinese, this situation is not allowed, they don't have the same meaning.
	\begin{enumerate}
		\item 03125643-a Tongan (of or relating to the island monarchy of Tonga or its people; "Tongan beaches" ),(汤加的-tangjiade,汤加语的-tangjiayude).
	\end{enumerate}
	Tongan can be people or place, we made the unified for this situation, the main part is retained, which is (汤加-tangjia).
	
	\textbf{\RNum {2}}  Processed some special lemmas, if the concept is verb, and the following structure exists, (使.. - shi.., ..于 - ..yu). Add "+" in them. For instance (..+于 - ..+yu). If the concept is adverb, and (..地 -..de) in lemma, the lemma becomes (..+地 - ..+de). The adjective is same as adverb, the lemma in adjective will become (..+的 - ..+baishaode).
	
	\textbf{\RNum {3}} One lemma can represent many meanings in PWN, we should translate into one Chinese word.
	\begin{enumerate}
		\item 10476928-n privateer (an officer or crew member of a privateer )
	\end{enumerate}
	In the gloss of 10476928, the word "officer" and "crew" are two different positions in Chinese, both of them cannot be confused, so we should retain only one.
	
	\textbf{\RNum {4}} There are many person's name in PWN, different translation mechanisms have different forms of translation. For instance (Henry\_Louis\_Aaron - 亨利路易斯亚伦-hengliluyisiyalun). So we make a standard: family name and first name are separated by dots. As follows:
	\begin{enumerate}
		\item 10807016-n Henry\_Louis\_Aaron (亨利·路易·斯亚伦-hengli·l
		
		uyi·siyalun)
	\end{enumerate}
	
	\textbf{\RNum {5}} Some lemmas consist of many words. In Chinese, this form does not exist. So we added lemmas for this kind concept, after adding:
	\begin{enumerate}
		\item 00196990-v  declaw (去除爪子-quchuzhuazi,去除-quchu)
	\end{enumerate}
	
	\subsubsection{The Hard Translation}
	The complexity of Chinese will lead to some concepts that cannot be used after translation. Dealing with them properly is what we do in the furture.
	\begin{enumerate}
		\item 01137415-v kneecap (用枪击穿膝盖骨-yongqiangjichuanxi
		
		gaigu)
		\vspace{0.5em}
		\item 10147849-n grocery\_boy (杂货店的男孩-zahuodiandenanhai)
	\end{enumerate}
	The strucure "verb+noun+verb+noun", "noun+noun" are very complex, in Chinese, they can not become a word. In the other hand, it is also not conducive to segmentation.
	
	\section{Compare Four Chinese Wordnets}
	In this section, we evaluate the effectiveness of MCW on three tasks including relatedness calculation, word similarity and word sense disambiguation, which mainly focusing on evaluation of the lemma's accuracy. We also compared coverage in four Chinese wordnets. Before this process, two standards were developed.

	\subsection{Creating standards}
	This section discusses how to create standards in Standard One and Standard Two.
	
	\subsubsection{Standard One}
	We assessed the lemma's accuracy by comparing the relatedness \cite{Yu2017Joint} of each Chinese lemma and its Chinese gloss, so we randomly chose 180 and 240 glosses in PWN \cite{Wang2014Building}. The following two criteria are met during the random selection process.

	\begin{enumerate}
		\item Randomly chose them from each layer of nodes in the PWN.
		
		\item Randomly chose them according to the proportion of 3:1:1:1 in nouns, verbs, adjectives and adverbs.
	\end{enumerate}
	
	In PWN, using the gloss of each concept is the best way to distinguish concepts. However, there are cases where the gloss is insufficient and inaccurate. For excample:\\
	\textbf{03079136-n} compact; compact\_car ("a small and econmical car")\\
	"small" and "econmical" cannot fully express the meaning of "compact", it should be described from many aspects, such as: the number of wheels, features, seats etc. \\
	\textbf{10142747-n} grandma ("the mother of your father or mother") \\
	In Chinese, the mother of father and mother are different concepts. It should be clear.
	
	For these problems, we manual tranlated 180 and 240 glosses which finished by 7 Chinese linguists, English linguists and translators. They are called C\_gloss180 and C\_gloss240. And available at \url{https://github.com/ToneLi/MCW_standard_one}

	\subsubsection{Standard Two}
	By calculating the similarity of lemmas in each Chinese wordnet can also be used to detect the accuracy of it. 
	So we adopted the 65 pairs of words published by Rubenstein and Goodenough \cite{Rubenstein1965Contextual} widely used in the world. And translated it into the corresponding Chinese word pairs by 5 translators, it is called C\_65. This is available at \url{https://github.com/ToneLi/MCW_standard_two}  

	\subsection{Coverage comparison}
	This section compared the number of concepts which are shown in Tabel 1. A total of three kinds of PWN were translated (wordnet1.6, wordnet2.0, wordnet3.0). In these four Chinese wordnets, COW only translated 35.96\%, there is an obvious shortage in coverage. SEW and CWN completed 95.78\% and 99.99\% respectively. MCW's proposal compensates for the lack of translation and the number of concepts exceeds other Chinese wordnets.

	\begin{table}[htbp]
		
		\centering  
		\caption{Comparison of the number of lemmas}
		\renewcommand\arraystretch{1.3}  
		\begin{tabular} { p{1cm}<{\centering} p{1cm}<{\centering} p{1cm}<{\centering} p{1cm}<{\centering} p{1cm}<{\centering}}
			\hline
			POS & SEW & CWN & COW & MCW  \\ 
			\hline   
			noun & 99581 &92440 &46229 & 118428 \\	
			verb & 24808 &20926 &13293 & 20041 \\
			adj & 31170 &30726 &18257 & 26160 \\
			adv & 6335 &5783  &2030 &5792 \\
			Total & 161894 &149875 &79809 &170421 \\
			\hline
			
		\end{tabular}
		
	\end{table}
	Lemmas' comparison is shown in Table 2. Compared four Chinese wordnets, COW has the least lemmas, MCW has the most lemmas which made up the lack of polysemous words in current Chinese wordnets.
	\subsection{Accuracy comparison}
	
	The accuracy of translation for synset lemmas is another important aspect of evaluation. We experimented on three tasks. 
	
	In relatedness calculation task, we calculated the relatedness \cite{Kleiman1980Sentence} between lemmas and glosses in every Chinese wordnet, Standard One and Word2vec-(skip-gram) were used for calculating the relatedness. Word2vec's parameters and training corpus are same as Section 3.
	In  word similarity task, Information Content(IC) \cite{Resnik1995Using} was used to calculate similarity \cite{Fellbaum1998Combining}. In word sense disambiguation task, we used four Chinese wordnets in Task5 about SemEval2007.

	\subsubsection{Relatedness Calculation}
	This section compares four Chinese wordnets with precision, recall and F-score \cite{Goutte2005A} by calculating the relatedness between lemmas and glosses.
	$$E=num[max_{1}(VE_{0-n})... \cup max_{i}(VE_{0-n})] \eqno{(2)} $$
	$$R=\frac{E}{ng}\eqno{(3)}$$
	We defined the formula for calculating recall. $VE_{0-n}$ means each lemma calculates the relatedness with all glosses.  $n$ means the number of glosses, it can be 180 or 240. $max_{i}$ means getting the most relevant gloss for each lemma, $i$ means the number of lemmas. For every concept, if its synset lemmas that correspond to the glosses are right, the concept is the right.
	$num$ means the number of right concept. $ng$ means the number of  C\_gloss180 and C\_gloss240, so it can be 180 or 240. 
	
	Next, we defined the precision, as follows:
	$$P=\frac{L}{S} \eqno{(4)}$$
	In this formula, $S$ means the total amount of lemma's gloss in each Chinese wordnet, it can also be the number of lemma. By the calculation, for every lemma, if its gloss that correspond to Standard One is right, the lemma is the right. $L$ means the number of right lemma.
	
	In the end, we defined the F-score:
	$$F=\frac{2*P*R}{P+R}\eqno{(5)}$$
	\begin{table}[htbp]
		\centering 
		\caption{Four Chinese Wordnets' R, P and F in C\_gloss180}
		\renewcommand\arraystretch{1.3} 
		\begin{tabular} {p{1cm}<{\centering}p{1cm}<{\centering}p{1cm}<{\centering}p{1cm}<{\centering}p{1cm}<{\centering}}
			\hline
			Wordnet & COW & CWN & SEW & MCW  \\ 
			\hline
			R &0.7833&0.6833 &0.7444 & 0.6555 \\
			
			P & 0.3369 &0.3846 &0.3101 & 0.4938 \\
			
			F & 0.4712 &0.4921 &0.4378 & 0.5633 \\
			\hline	
		\end{tabular}
		
	\end{table}
	
	\begin{table}[htbp]
		\centering  
		\caption{Four Chinese Wordnets' R, P and F in C\_gloss240}
		\renewcommand\arraystretch{1.3} 
		\begin{tabular} {p{1cm}<{\centering}p{1cm}<{\centering}p{1cm}<{\centering}p{1cm}<{\centering}p{1cm}<{\centering}}
			\hline
			Wordnet & COW & CWN & SEW & MCW  \\ 
			\hline   
			
			R &0.8000&0.7042 &0.7542 & 0.6958 \\
			
			P & 0.3183 &0.3716 &0.3098 & 0.5137 \\
			
			F & 0.4554 &0.4865 &0.4391 & 0.5910 \\
			\hline	
		\end{tabular}
		
	\end{table}
	The results of the relatedness calculation are indicated in Table 3 and Table 4. Four Chinese wordnets show lower accuarcy on same sets, the reason is as follows:

\begin{enumerate}
	\item  In Standard One, we think each lemma and its gloss are completely related, relatedness is 1. However semantics is very complicated in Chinese, machines that rely on statistics to accomplish tasks do not understand the semantics correctly.
	
	\item Different research teams have different criteria in the judgment of synonyms. Such as, (COW-00364221-r: 确定的-qudingde, 不动地-budongde). COW researchers think these two words belong to the same concept, they are synonym. However, in SEW, they are different concepts.
\end{enumerate}

	After comparison, MCW has a high accuracy, because of its construction focused on the translation of synonyms. COW has high recall, but has lower precision, it shows that more glosses have been recalled through calculations, editors have a shortage of synonyms, just focus on the distinction between core synsets. SEW has the lowest precision, this is because editors are mainly focusing on the quantity of wordnet, ignoring many details. Such as, in (609456014-n), there are four different meanings. (工头-gongtou, 母牛-muniu, 浮雕-fudiao, 瘤-liu). They should be synonymous. Another situation, lemma consists of lemma and its explanation, each lemma has a unique meaning, excessive explanation will lead to a decline in the accuracy of SEW. (609702631-n: 老兄(对男人的昵称)-laoxiong(nickname for man)), CWN's score is lower, this is because it has the same situation as SEW, in addition, there are many vocabularies that are different from the mainland and not universal.
	\subsubsection{Word Similarity}
	In this section, we used the method of conceptual similarity to evaluate four Chinese wordnets, and experimented on C\_65.
	
	In PWN, the method of conceptual similarity depends on two sub-calculation methods: the calculation method of conceptual InformationContent (IC), the calculation method of conceptual similarity (Sim) based on IC. We combined these two methods.
	
	\cite{Zhou2008A}'s IC calculation method is used, this method takes into account the depth of the concept, improves the calculation accuracy.
	$$IC_{zhou}(C)=k(1-\frac{\log(|hypo(C)|+1)}{\log(max\_nodes)})+
	(1-k)(\frac{\log(|depth(C)|+1)}{\log(max\_depth)})\eqno{(6)}$$
	In this formula, $max\_nodes$ represents the maximum number of nodes on the ontology, $|hypo(C)|$ represents the number of all hypogyny nodes of concept $C$ in the ontology hierarchy. $depth(C)$ represents the depth of concept $C$ in the ontology hierarchy.  $|max\_depth|$ represents the maximum depth in the ontology hierarchy, $k$ is the adjustment weight factor.
	Lin's Sim calculation method is used:
	$$Sim_{lin}(C_1,C_2)=\frac{2*IC(LCS(C_1,C_2))}{IC(C_1)+IC(C_2)}\eqno{(7)}$$
	In this formula, $C_1$ and $C_2$ represent the concept, $LCS(C_1,C_2)$ represent the closest public parent node of  $C_1$ and $C_2$. $IC(C)$ represent the  InformationContent of concept $C$.
	
	Based on the above, we defined the formula for solving the conceptual similarity.
	$$msim=max\{Sim_{lin}[id(lma1)_i,id(lma2)_j]\} \eqno{(8)}$$
	$$SIM=Spearman(msim,man\_sim)\eqno{(9)}$$
	In PWN, ids are used to represent concepts. We filtered lemmas which same as C\_65 in every Chinese wordnet, then found the ids in PWN for those lemmas. Every lemma has one or more than one ids in those four wordnets, $id(lma)_i$ represents the ids for every lemma, $i$ and $j$ both are 65, due to the above situation, we calculated the maximum IC for each pair lemmas' ids. Finally using spearman to obtain the similarity between these results($msim$) and manually revised similarity($man\_sim$).
	\begin{table}[htbp]
		\centering  
		\caption{Four Chinese wordnets' conceptual similarity in C\_65}
		\renewcommand\arraystretch{1.3} 
		\begin{tabular} {p{1cm}<{\centering} p{1cm}<{\centering}p{1cm}<{\centering}p{1cm}<{\centering}p{1cm}<{\centering}}
			\hline
			Wordnet & COW & CWN & SEW & MCW  \\ 
			\hline   
			
			similarity &0.8351&0.7868 &0.8329 &  0.8739\\
			\hline 
		\end{tabular}
	\end{table}
	
	We derived the result from the perspective of similarity, it was shown in Table 5. CWN has the lowest accuracy, this is becauce every lemma has too much wrong ids, lemmas' semantic information is not the same as concept's semantic information. COW and SEW has roughly same result . MCW focus on the semantic information between lemmas and concepts, so its similarity is high.
	\subsubsection{Word Sense Disambiguation}
	The multilingual Chinese-English lexical sample task at SemEval-2007 provides a framwork to evaluate Chinese word sense disambiguation. In this section, we evaluated the Chinese wordnet in SemEval-2007 (task5) which about Chinese word sense disambiguation.

	\begin{figure}[htbp]  
		\centering
		\includegraphics[height=5.9cm,width=0.25\textheight]{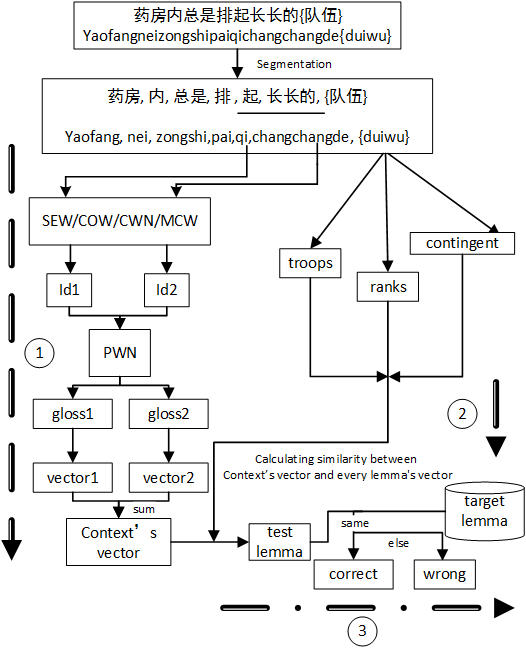}
		\caption{The flow of Chinese word sense disambiguation}\label{fig-3}  
	\end{figure}
	In this task, we experimented on its test data, together 36 word-types about Chinese ambiguous words: 17 nouns and 19 verbs, a total of 935 Chinese sentences, every sentence has one Chinese ambiguous word  that need to be disambiguated. These ambiguous words contain one or more lemmas. In Fig.3, the ambiguous word is 队伍 (duiwu). Its lemmas are troops, ranks and contingent. They represent different meanings. In our word sense disambiguation algorithm, there are four steps, as follows:
	
	(1) Preprocessing. In order to generate the correct meaning of ambiguous words, we need to segment the sentences and remove the stop words.
	
	(2) Establish a contextual environment. It is shown in Fig.3-marker 1. In our algorithm, we used words which window is 2 as a contextual environment. In Fig.3, the context is 起 (qi) and 长长的 (changchangde). By searching in Chinese wordnet, the word in context can find its id which same as PWN. Every id in PWN has the gloss to define it. We used gloss to represent the context. And then, we used the vector space model to characterize the glosses, these vector add up to context's vector.

	(3)  Word sense disambiguation. It is shown in Fig.3-marker 2,3. Calculating similarity between Context vector and every lemma's vector. By calculating the maximum, we can get the word meaning of disambiguation. If it is same as the target, it's correct, else is wrong.
	
	(4) Verification. Two kinds of precision are evaluated. One is micro-average:
	$$ P_{mir}=\sum_{i=1}^Nm_i / \sum_{i=1}^Nn_i \eqno{(10)}$$
	$N$ is the number of all word-types (36), 	$m_{i}$ is the number of labeled correctly to one specific word-type and $n_{i}$ is the number of all instances for this word-type.
	
	The other is macro-average:
	$$ P_{mar}=\sum_{i=1}^Np_i / N, p_{i}=m_{i}/n_{i}\eqno{(11)}$$ 
	
	The results as follows:

	\begin{table}[htbp]
		\centering  
		\caption{The scores of all Chinese wordnets}
		\renewcommand\arraystretch{1.3}  
		\begin{tabular} {p{2cm}<{\centering} p{2cm}<{\centering} p{3cm}<{\centering}}
			\hline
			Wordnet & Micro-average & Macro-average\\ 
			\hline   	
			COW &0.5432&0.5632  \\
			CWN & 0.4501 &0.4673 \\
			SEW & 0.4154 &0.4365  \\
			MCW & 0.6703 &0.7132 \\
			baseline & 0.4053 &0.4618\\
			\hline	
		\end{tabular}
	\end{table}
	
	As shown in Table 6, our MCW outperforms other Chinese wordnet, on both two precisions. This indicates that, by utilizing machine Learning and manual correction, MCW can better describe Chinese ontological network and indicate the relationship between entities. We choose the context window as 2 as the semantic environment. Some words in the window cannot find the corresponding glossed in PWN, because these wordnets' coverage is insufficient. This leads to lower overall accuracy. On the other hand, the window word cannot represent complete context information, in the future, we will use a variety of tools such as dependency syntax analysis to represent it. These Chinese wordnet significantly outperform baseline, this also proves the validity of Chinese wordnet for word sense disambiguation.

	\section{Conclusions and Future Work}
	
	In this paper, we proposed a Chinese novel wordnet (MCW) and used different theories and tools to make it. Specifically, the machine learning was introduced into the building process for the first time. We evaluated MCW on relatedness calculation, word similarity and word sense disambiguation, and results show the advantages of MCW. We also analyzed the coverage of several Chinese wordnets, which confirmed MCW is the biggest Chinese wordnet currently. Chinese have complex semantic structure that is difficult to process, although we carefully constructed MCW, there still some wrong lemmas in it. This leaves us space for more improvements, and expands a bigger Chinese wordnet. Further more, we will also produce multilingual wordnets one after another.  It will be applied to widely field.
		
		
		\begin{acks}
		This research is funded by the Digital Humanities Center of Qufu Normal University, and by Science and Technology Commission of Shanghai Municipality (No. 16511102702) and by Xiaoi Research, by Shanghai Municipal Commission of Economy and Information Under Grant Project No. 201602024 and by the Science and Technology Commission of Shanghai Municipality (No.15PJ1401700).
		\end{acks}
		
		\bibliographystyle{ACM-Reference-Format}
		\bibliography{sample-base}
		
		\appendix

	\end{CJK*}
\end{document}